\documentclass[letterpaper]{article} 
\usepackage{aaai24}  
\usepackage{times}  
\usepackage{helvet}  
\usepackage{courier}  
\usepackage[hyphens]{url}  
\usepackage{graphicx} 
\usepackage{natbib}  
\usepackage{caption} 
\usepackage{algorithm}
\usepackage{algorithmic}
\usepackage{newunicodechar}

\usepackage{amsmath}
\usepackage{amsfonts}
\usepackage{booktabs}
\usepackage{array}
\usepackage{pifont}
\usepackage{booktabs}
\usepackage[table]{xcolor}
\definecolor{passgreen}{HTML}{C6EFCE}
\definecolor{failred}{HTML}{FFC7CE}
\newcommand{\pass}{\cellcolor{passgreen}\ding{51}}

\usepackage{newfloat}
\usepackage{listings}
\DeclareCaptionStyle{ruled}{labelfont=normalfont,labelsep=colon,strut=off} 
\floatstyle{ruled}
\newfloat{listing}{tb}{lst}{}
\floatname{listing}{Listing}
\title{LLM-QUBO: An End-to-End Framework for Automated QUBO Transformation from Natural Language Problem Descriptions}
\author{
    Huixiang Zhang\textsuperscript{\rm 1},
    Mahzabeen Emu\textsuperscript{\rm 2},
    Salimur Choudhury\textsuperscript{\rm 3}
}
\affiliations{
    \textsuperscript{\rm 1}Department of Computer Science, Lakehead University\\
    955 Oliver Road, Thunder Bay, Ontario, Canada P7B 5E1\\
    \textsuperscript{\rm 2}Quantum Communications and Computing Research Center, Department of Electrical and Computer Engineering, Memorial University of Newfoundland\\
    P.O. Box 4200, St. John's, NL A1C 5S7, Canada\\
    \textsuperscript{\rm 3}Queen's School of Computing, Queen's University\\
    99 University Avenue, Kingston, Ontario, Canada K7L 3N6\\
    hzhan102@lakeheadu.ca, memu@mun.ca, S.choudhury@queensu.ca
}

\usepackage{bibentry}

\begin{document}

\maketitle

\begin{abstract}
Quantum annealing offers a promising paradigm for solving NP-hard combinatorial optimization problems, but its practical application is severely hindered by two challenges: the complex, manual process of translating problem descriptions into the requisite Quadratic Unconstrained Binary Optimization (QUBO) format and the scalability limitations of current quantum hardware. To address these obstacles, we propose a novel end-to-end framework, LLM-QUBO, that automates this entire formulation-to-solution pipeline. Our system leverages a Large Language Model (LLM) to parse natural language, automatically generating a structured mathematical representation. To overcome hardware limitations, we integrate a hybrid quantum-classical Benders' decomposition method. This approach partitions the problem, compiling the combinatorial complex master problem into a compact QUBO format, while delegating linearly structured sub-problems to classical solvers. The correctness of the generated QUBO and the scalability of the hybrid approach are validated using classical solvers, establishing a robust performance baseline and demonstrating the framework's readiness for quantum hardware. Our primary contribution is a synergistic computing paradigm that bridges classical AI and quantum computing, addressing key challenges in the practical application of optimization problem. This automated workflow significantly reduces the barrier to entry, providing a viable pathway to transform quantum devices into accessible accelerators for large-scale, real-world optimization challenges.
\end{abstract}

\section{Introduction}
Combinatorial optimization problems constitute a class of computational challenges of significant value in modern science, finance, logistics, and engineering. For canonical problems such as the Traveling Salesman Problem or portfolio optimization, the solution space grows exponentially with the problem size, rendering traditional computational methods intractable for large-scale instances. Quantum computing offers a promising alternative paradigm. Its core premise is that by encoding a problem's objective function as the energy function (Hamiltonian) of a physical system, a quantum annealer can naturally evolve to its lowest energy state, which corresponds to an optimal or near-optimal solution. The potential for quantum speedup is a primary driver for research, although current adiabatic quantum computing devices could not be proven to be faster than classical computing resources yet\cite{mucke2023}. This prospect has catalyzed extensive exploration in various applications, including financial asset allocation, biomedical research, and logistics network design \cite{morapakula2025, oliveira2018, malviya2023}.

However, a stark reality on the path to quantum advantage is that we are in the era of noisy intermediate-scale quantum computing \cite{holliday2025}. Current quantum processors remain severely constrained in qubit count, coherence times, and connectivity \cite{morapakula2025}. Consequently, the most viable path to realizing the potential of quantum computing in the foreseeable future lies not in purely quantum algorithms but in building Hybrid Quantum-Classical (HQC) systems \cite{zhao2022}. This paradigm advocates for a strategic division of labor: classical High-Performance Computing (HPC) systems handle tasks at which they excel, such as data pre- and post-processing, control flow, and numerically intensive computations, while quantum annealer function as specialized hardware accelerators or coprocessors, focused on solving the most computationally intensive and combinatorial complex parts of the problem. This synergistic model aims to merge the robustness of classical computation with the exploratory power of quantum computation, forming a powerful platform whose capabilities extend beyond those of any single paradigm alone \cite{zhao2025}.

The central idea of this paper is that while the HQC architecture charts a path toward practical quantum computing, a fundamental obstacle, which we term the Formulation Bottleneck, blocks the realization of its potential. In the quantum annealing paradigm, any problem must be converted to the Quadratic Unconstrained Binary Optimization (QUBO) format as a middle layer between classical optimization problems and quantum hardware, minimizing an objective function of the form $y=x^{T}Qx$ \cite{glover2022}. However, translating real-world problems into this representation is an exceptionally complex, expert-dependent, and error-prone task. It involves specialized skills such as constraint to penalty term translation, manual penalty coefficient tuning, and selecting appropriate QUBO precision, which not only raises the barrier to entry but also complicates integration into existing HPC workflows \cite{ayodele2022, volpe2024}. The essence of this challenge is the lack of a high-level abstraction layer, analogous to the compilers in classical computing that automatically translate high-level code into machine instructions \cite{zaman2022}. To bridge this gap, our framework provides this necessary layer by creating an end-to-end pipeline that automates the transformation from a high-level problem description into an optimized QUBO matrix. This automated pipeline fundamentally transforms the skill set required to apply quantum computing. Our framework encapsulates the underlying complexity, shifting the primary challenge for a user from ``How do I construct a valid QUBO?'' to ``How do I precisely formulate my optimization problem?'' This skill is the root of classical operations research, not quantum computing. This reduction in the specialized knowledge barrier is fundamental to fostering a robust ecosystem of quantum optimization tools and accelerating their adoption across science and industry.

This paper presents a comprehensive framework that achieves this goal through the following key contributions.

\begin{itemize}
\item \textbf{An LLM-driven compiler for end-to-end QUBO formulation.} We propose and implement a novel framework that leverages an LLM to automate the transformation from a high-level problem description into a quantum-ready QUBO matrix, significantly lowering the barrier to entry for quantum optimization.

\item \textbf{A validated HQC workflow for scalability.} We integrate Benders' Decomposition into our automated pipeline to tackle large-scale problems. This hybrid approach partitions a problem into a QUBO master problem suitable for quantum annealers, and subproblems solved by classical HPC resources, providing a viable pathway to solve problems that exceed the capacity of standalone quantum processors.

\item \textbf{A novel integration of AI, HPC and quantum.} We demonstrate for the first time a seamless workflow that combines a generative AI compiler with a hybrid quantum decomposition solver, establishing a new paradigm for automated problem solving.

\end{itemize}

\section{Related Work}
We review related work by deconstructing the pipeline for transforming a combinatorial optimization problem into its QUBO representation. This pipeline includes several critical stages: problem formulation from natural language, binarization of variables, conversion of constraints into penalty terms, and the final generation of the QUBO matrix.

\subsection{Optimization Problems Formulation From Natural Language Description} 
The task of converting a problem described in natural language into a formal mathematical model is a significant challenge that has traditionally required deep domain expertise. The emergence of LLM has catalyzed a field of research known as auto formulation, aimed at automating this process \cite{huang2025}.
Foundational efforts, such as the NL4Opt Competition, spurred the development of learning-based methods by structuring the task into subproblems like entity recognition and logical form generation \cite{ramamonjison2021, ramamonjison2022}. This work highlighted core challenges, including handling unstructured inputs and the need for models to generalize across different problem domains. Subsequent research has benchmarked prominent LLMs on this task, where large models such as GPT-4 established new performance levels without requiring the input of entities of prior baselines. The same work also introduced the LM4OPT framework, revealing a persistent capability gap between large models and smaller, fine-tuned ones when processing lengthy and complex problem contexts \cite{ahmed2024}.
Other research has explored different frameworks and methodologies. For example, Jiang et al. introduced the LLMOPT framework, which employs a learning-based method to fine-tune LLMs to automate MILP formulation, showing a significant increase in the precision of the solution \cite{jiang2025}. In a different paradigm, OptiChat was developed as an LLM-assisted interactive dialogue system. Instead of formulating problems from scratch, it helps practitioners interpret and query existing optimization models by augmenting the LLM with targeted code generation to ensure trustworthy responses \cite{chen2025}.

\subsection{QUBO Matrix Generation}
The AutoQUBO framework, introduced by Moraglio et al. and enhanced by Pauckert et al., represents a contribution to converting combinatorial optimization problems into the QUBO format. The framework accepts cost and constraint functions defined in a high-level programming language such as Python. To derive the QUBO coefficients, it employs a data-driven interpolation method that samples the problem description using a special selection of binary input vectors, specifically, the all-zeros vector, one-hot encoded vectors, and two-hot encoded vectors. This process systematically determines the constant, linear, and quadratic coefficients of the expression QUBO. Subsequently, AutoQUBO v2 can construct separate QUBO matrices for the cost and constraint functions, automatically estimate a valid penalty weight based on the cost matrix, and finally combine them to produce the final QUBO matrix for the problem.  \cite{moraglio2022, pauckert2023}.

However, the framework's approach to non-binary variables, particularly continuous ones, has a limitation based on direct binarization. For a problem with continuous variables, such as an Uncapacitated Facility Location Problem (UFLP) with $N$ facilities and $M$ customers, if each continuous assignment variable is encoded with $K$ binary bits, the total number of variables in the QUBO model scales to $N+(N \times M \times K)$. Consequently, the size of the resulting QUBO matrix grows quadratically as $(N+(N \times M \times K))^2$. This explosive growth in dimensionality, especially with a large number of customers $M$ or required precision $K$, poses a fundamental scalability bottleneck, limiting the framework's applicability for solving complex combinatorial problems that involve continuous variables.

\subsection{Research Gap}
While the above works have made significant progress in automating the initial step from natural language to a structured model like MILP, a critical bottleneck remains in the subsequent stage: the translation from a classical optimization model into a hardware-compatible QUBO format. This process is not only complex and error-prone, but also crucial for leveraging quantum hardware. To bridge this research gap, our work is driven by two central questions.

\begin{itemize}
\item How can we leverage LLMs to reliably automate the expert-driven, rule-intensive conversion of a structured MILP into a correct and efficient QUBO representation?
\item How can this automated QUBO generation be integrated into a scalable, hybrid quantum-classical framework to overcome the limitations of near-term quantum devices and solve large-scale problems?
\end{itemize}

\section{Framework Design}

This section introduces the architecture of our framework, which leverages a LLM to automate the complex pipeline from a natural language problem description to a quantum-ready solution. The framework is designed to address the primary bottleneck in applying quantum computing: the manual, expert-driven conversion of optimization problems into the QUBO format.

\begin{figure*}[h] 
\centering
\includegraphics[width=1\linewidth]{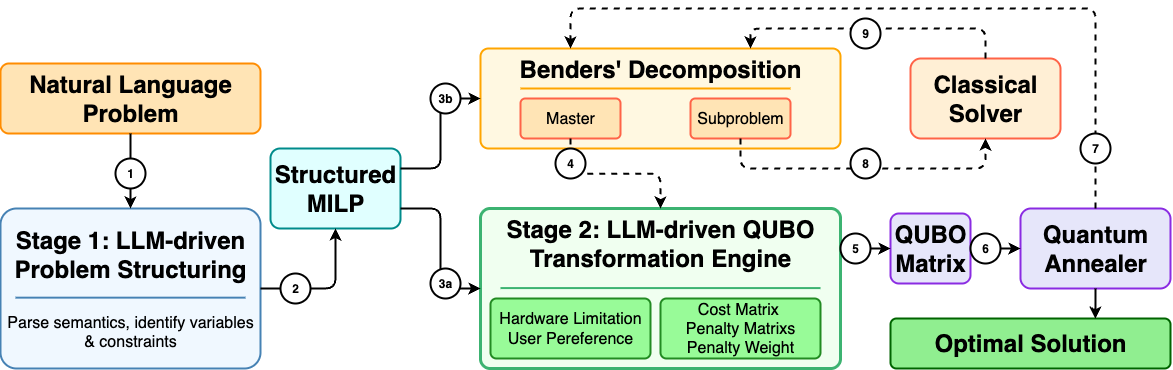}
\caption{Framework Overview} 
\label{fig:overview}
\end{figure*}

\subsection{Overall Architecture and Workflow}
The architecture of our proposed framework is illustrated in Figure 1. It is an end-to-end pipeline that begins with a natural language problem description as its primary input. 

The initial step of this pipeline is Stage 1: LLM-driven Problem Structuring. In this stage, the framework uses an LLM to translate the unstructured problem description into a formal mathematical model. The process involves guiding the LLM to parse the input text and identify five categories: sets, over which indices are defined; parameters, the constants of the problem; decision variables, including their types (e.g., binary, continuous); the objective function with its optimization direction; and the constraints. The LLM then synthesizes these extracted components into a structured MILP, which serves as the standardized input for subsequent transformation stages. The accurate classification of variable types at this stage is critical, as it directly informs the logic of the QUBO conversion process. From the Structured MILP, the framework supports two distinct operational workflows.

For smaller problems, a direct conversion path (Path 3a) is taken, where the MILP is fed into Stage 2: LLM-driven QUBO Transformation Engine. For large-scale problems that would exceed the capacity of current quantum hardware, the framework employs a hybrid decomposition strategy (path 3b). In this workflow, the MILP is first partitioned by a Benders' Decomposition module into a combinatorial Master Problem and a linearly structured Subproblem. The Master Problem is then sent to the Stage 2 engine for QUBO conversion (Path 4), while the Subproblem is handled by a Classical Solver.

The final QUBO matrix is dispatched to a Quantum Annealer to find an optimal solution. In the hybrid workflow, an iterative refinement process begins. The solutions from the quantum and classical solvers are used to generate Benders' cuts, which are fed back to the Benders' Decomposition module to refine the Master Problem (Path 9). This loop continues until a predefined convergence criterion is satisfied, which yields the final optimal solution to the original large-scale problem.

\subsection{QUBO Transformation Engine}
The second and core stage of our framework is the automated conversion of the structured MILP into a QUBO format. We address this challenge by using the LLM as an expert rule-based conversion engine through a process of structured prompt engineering.

Our methodology guides the LLM to systematically deconstruct the input MILP and resynthesize it into a structured Python class suitable for QUBO generation tools like AutoQUBO. This class explicitly separates the objective function from the constraint penalty terms. The LLM is instructed to adhere to the following key conversion principles.

\begin{itemize}
    \item \textbf{Common Constraint Conversion:} The LLM handles standard constraints by converting them into quadratic penalty terms. Equality constraints of the form $LHS = RHS$ are directly formulated as a penalty $(LHS - RHS)^2$. For inequalities such as $LHS \le RHS$, the process involves first introducing a binarized slack variable, $s$, to form an equivalent equality, $LHS + s = RHS$. This is then converted into its corresponding quadratic penalty, $(LHS + s - RHS)^2$. To conserve resources, the number of binary bits for the slack variable $s$ is minimized based on the constraint parameters, following a ``just enough'' precision rule.

    \item \textbf{Specialized Structure Recognition:} Beyond general rules, our prompt engineering also equips the LLM to identify special constraint structures that allow for more efficient QUBO formulations. A notable example is the pairwise exclusion constraint $x_i + x_j \le 1$, where $x_i$ and $x_j$ are binary variables. Instead of mechanically applying the slack variable method, the LLM is guided to recognize that this constraint is violated only when $x_i=1$ and $x_j=1$. This violation is captured by the more compact quadratic penalty term $P \cdot x_i x_j$, where $P$ is a sufficiently large penalty coefficient. This specialized handling avoids the introduction of unnecessary slack variables, resulting in a more efficient final QUBO model.
\end{itemize}

To ensure that the resulting QUBO model is compatible with the target hardware, non-binary variables must be converted into a binary representation. Our framework manages this binarization process while being aware of physical hardware limitations.

An integer variable $y$ within a range $[0, U]$ is binarized using a standard binary expansion. It is replaced by a sum of $k$ new binary variables, $z_0, z_1, \dots, z_{k-1}$:
$$y = \sum_{i=0}^{k-1} 2^i z_i$$
The precision $k$ is the minimum number of bits required to represent the upper bound $U$, calculated as $k = \lceil\log_2(U+1)\rceil$. A critical aspect of our framework is its awareness of the quantum annealer capacity. For example, assume that a problem has 15 native binary variables and one integer variable and that the target quantum annealer can only accept a QUBO matrix of up to $24 \times 24$. This imposes a hard limit on the total number of variables: $15 + k \le 24$, which implies that the binarization precision for the integer variable cannot exceed $k=9$ bits. Our structured prompts instruct the LLM to calculate and adhere to such hardware-aware precision limits for all non-binary variables, ensuring that the final QUBO model is physically solvable.

The final QUBO matrix is the original cost matrix and a weighted sum of all penalty matrix derived from the constraints. The general form is:
$$ \text{QUBO Matrix} = \text{Cost Matrix} + \sum_{j} P_j \cdot \text{Penalty Matrix}_j $$
where $P_j$ is the penalty coefficient for the $j$-th constraint.

Although existing tools often use a single penalty weight P, this is not good enough since real-world constraints may have varying importance. Our LLM-driven approach opens the path for inferring the semantic priority of each constraint from the original problem description to assign distinct weights ($P_j$). For example, the LLM can assign a higher penalty to a critical physical capacity constraint than to a less rigid budget constraint. This intelligent weighting leads to a better-conditioned QUBO model that guides the solver more effectively towards a valid and optimal solution.


\subsection{Application: Solving Large-Scale Problems via Hybrid Benders' Decomposition}

While our LLM-driven conversion engine can process any suitable MILP, converting a large-scale problem into a single, monolithic QUBO often results in a model that is too large and complex to be solved effectively. This scalability challenge, which we demonstrate empirically in Section 4, motivates the use of a decomposition strategy. To this end, we apply our conversion methodology within a hybrid Benders' decomposition framework to tackle large-scale optimization problems.

This strategy partitions a source MILP into more manageable components. Our framework assumes the MILP can be expressed in the following general form:
\begin{equation}
\begin{array}{ll}
\text{minimize} & c^T x + f^T y \\
\text{subject to} & Ax + By \leq b \\
& x \in \mathbb{R}_+^n, \quad y \in \{0, 1\}^p
\end{array}
\end{equation}
Here, $y$ represents the vector of $p$ binary decision variables that capture the combinatorial complexity of the problem, while $x$ is the vector of $n$ continuous variables. The Benders' decomposition method reformulates this problem into two smaller, linked problems:

\begin{enumerate}
    \item \textbf{The Master Problem (MP):} An integer program involving only the binary variables $y$. It approximates the impact of the continuous variables through a set of constraints known as Benders' cuts, which are added iteratively.
    \item \textbf{The Subproblem (SP):} A linear program that solves for the continuous variables $x$, assuming the values of the binary variables $y$ are fixed.
\end{enumerate}

The critical integration point for our core contribution occurs in the master problem. The isolated MP, now a pure binary optimization problem, becomes the direct input for our LLM-driven conversion engine described in Stage 2. The LLM applies the rules-based methodology to transform the MP into a compact, hardware-aware QUBO.

The overall hybrid workflow proceeds iteratively. In each iteration, the MP-QUBO is solved to propose a new set of binary decisions. The SP is then solved with these decisions fixed. The solution to the SP is used to generate a new Benders' cut that is added to the MP for the next iteration, progressively refining the solution space. As established in our framework's introduction, the resulting QUBO is validated in our experiments using classical solvers. This approach confirms the correctness of the decomposition and conversion, benchmarking the structure's performance and verifying the quantum hardware compatibility of the QUBO formed master problem for future implementation on a physical quantum annealer.

\section{Experiment and Result Analysis}

Recent work demonstrates that LLM can accurately translate natural language descriptions of optimization problems into formal MILP models, with frameworks such as LLMOPT achieving high accuracy on standard benchmarks. Given this established capability, our work begins with the assumption of an accurate pre-existing MILP model. Therefore, our experimental focus is on the automated, high-quality conversion of these MILPs into QUBO formulations.

\subsection{LLM Automatic Modeling}

The LLM automatic modeling tasks were performed using a Qwen3-8B model, locally deployed on a single NVIDIA A40 GPU. The primary objective of our automated conversion framework is to produce QUBO models that are not only syntactically valid but also semantically equivalent to the source MILP. We assess the quality of the conversion based on three criteria: (1) adherence to the standardized input format required by tools like AutoQUBO, which separates the cost function from constraint penalty terms; (2) the correct application of binarization strategies for non-binary decision variables and the introduction of slack variables for inequality constraints; and (3) the mathematical correctness of the penalty function formulation for each constraint. 

Our analysis, summarized in Table 1, evaluates the conversion of nine classical optimization problems. The results indicate that, while many problems can be converted successfully, significant challenges arise from complex constraint structures, often leading to incorrect penalty formulations. To illustrate these findings, we analyze one correct conversion and one incorrect conversion in detail.

\begin{table*}[t!]
\centering
\footnotesize
\caption{Analysis of automated MILP-to-QUBO conversion correctness for nine classical optimization problems.}
\label{tab:conversion_analysis_final}
\begin{tabular}{@{} >{\raggedright\arraybackslash}m{2.3cm} >{\centering\arraybackslash}m{1.2cm} >{\raggedright\arraybackslash}m{2.7cm} >{\raggedright\arraybackslash}m{4.4cm} >{\raggedright\arraybackslash}m{2.3cm} >{\raggedright\arraybackslash}m{2.2cm}}
\toprule
\textbf{Problem} & \textbf{Structure} & \textbf{Decision Variable} & \textbf{Constraint Type} & \textbf{Encoding Strategy} & \textbf{Penalty Correctness} \\
\midrule

Traveling Salesman Problem & \pass & Binary \texttt{x}, Continuous \texttt{u} & Equality (Degree) \& Inequality (Subtour Elimination) & \cellcolor{passgreen}Dynamic bits for \texttt{u}. & \cellcolor{failred}Has high order item.\\
\addlinespace[1pt]

Weighted Max-Satisfiability & \pass & Binary \texttt{x}, \texttt{z} & Inequality (Logical Clause Association) & \cellcolor{passgreen}Dynamic bits for slack \texttt{s}. & \pass \\
\addlinespace[1pt]

Vehicle Routing Problem & \pass & Binary \texttt{x} & Equality (Flow/Visit) \& Inequality (Capacity) & \cellcolor{passgreen}Dynamic bits for slack \texttt{s}. & \pass \\
\addlinespace[1pt]

Portfolio Optimization & \pass & Continuous \texttt{w} & Equality (Budget) \& Inequality (Minimum Investment & \cellcolor{failred}Not performed. & \cellcolor{failred}Incorrect penalty function. \\
\addlinespace[1pt]

Maximum Clique & \pass & Binary \texttt{x} & Inequality (Pairwise Node Exclusion) & \cellcolor{passgreen}N/A & \pass \\
\addlinespace[1pt]

Maximum Independent Set & \pass & Binary \texttt{x} & Inequality (Pairwise Node Exclusion) & \cellcolor{passgreen}N/A & \pass \\
\addlinespace[1pt]

Logistics Network Design & \pass & Binary \texttt{y}, Continuous \texttt{x} & Equality (Demand) \& Inequality (Performance Metrics) & \cellcolor{passgreen}Dynamic bits for flow \texttt{x}. & \pass \\
\addlinespace[1pt]

Knapsack Problem & \pass & Binary \texttt{x} & Inequality (Single Capacity) \texttt{s}. & \cellcolor{passgreen}Dynamic bits for slack \texttt{s}. & \pass \\
\addlinespace[1pt]

Capacitated Facility Location & \pass & Binary \texttt{x}, \texttt{y} & Multiple Types (Coverage, Metrics, Capacity) & \cellcolor{passgreen}Dynamic bits for slack \texttt{s}. & \pass \\

\bottomrule
\end{tabular}
\end{table*}

\subsubsection{Correct Conversion Example: Capacitated Facility Location Problem}

CFLP conversion demonstrates robust and correct handling of multiple and diverse constraint types. This problem includes equality, inequality, and indicator logic constraints, all of which were transformed correctly.

\begin{itemize}
\item The customer service equality constraint, $\sum_{i} x_{ij} = 1$, was correctly penalized using the standard quadratic form $(\sum_{i} x_{ij} - 1)^2$.

\item The indicator logic constraint, $x_{ij} \le y_i$, which links the binary variables for customer assignment ($x_{ij}$) and facility opening ($y_i$), was efficiently converted into the compact quadratic penalty term $x_{ij}(1 - y_i)$.

\item Most importantly, the capacity inequality constraint, $\sum_{j} d_j x_{ij} \le C_i y_i$, was correctly transformed into an equality by introducing a binarized integer slack variable $s_i$. This led to the mathematically sound penalty function $(\sum_{j} d_j x_{ij} + s_i - C_i y_i)^2$.
\end{itemize}

The successful conversion of the multifaceted CFLP highlights that a rules-based, systematic approach can produce correct and high-quality QUBO formulations, provided that established principles for handling different constraint types are strictly followed.

\subsubsection{Incorrect Conversion Example: Traveling Salesman Problem (TSP)}

The TSP conversion successfully binarized the continuous auxiliary variables $u_i$ from the Miller-Tucker-Zemlin (MTZ) sub-tour elimination constraints. However, it fails on the third criterion regarding penalty correctness. The MTZ inequality is formulated as $u_i - u_j + n \cdot x_{ij} \le n-1$.

The implemented penalty term for this constraint was $\max(0, u_i - u_j + n \cdot x_{ij} - (n-1))^2$. Although this function correctly identifies violations, its expansion after substituting the binarized representation of $u_i$ and $u_j$ results in a polynomial of an order greater than two. This violates the quadratic nature of QUBO and creates a model that is incompatible with standard solvers.
The correct approach requires the introduction of a binarized slack variable $s_{ij}$ to form an equality $u_i - u_j + n \cdot x_{ij} + s_{ij} = n-1$. This equality can then be correctly penalized with a standard quadratic term $(u_i - u_j + n \cdot x_{ij} + s_{ij} - (n-1))^2$.

\subsection{Scalable Hybrid Computation}
To evaluate the performance and scalability of our proposed HQC framework, we used the Capacitated Facility Location Problem (CFLP) as our testbed. This NP-hard problem is ideal for Benders' decomposition due to its inherent structure, ensuring a fair comparison across methodologies. All experiments were conducted on an Apple M4 Pro CPU with Gurobi 12.0.3, using problem instances from the OR-Library benchmark suite for reproducibility.
\subsubsection{Benchmarking Methodology}
We compare the performance of our framework against two critical baselines, representing the state-of-the-art classical approach and the naive quantum-inspired approach, respectively.

\begin{itemize}
\item Method 1: Direct MILP solution (Gurobi). The complete CFLP is formulated as a standard Mixed-Integer Linear Program and solved directly using Gurobi. This serves as our baseline for both solution quality and classical performance.
\item Method 2: The entire QUBO matrix solution. The entire CFLP, including binarized representations of all variables and constraints, is converted into a single large QUBO matrix using our framework. This large QUBO matrix is then solved using Gurobi's quadratic programming capabilities. This method represents the performance of a direct, nondecomposed QUBO approach.
\item Method 3: Hybrid Benders Decomposition (Our Approach). The CFLP is partitioned according to our framework's logic. The master problem, which contains the binary facilities opening variables, is formulated as a QUBO. The subproblem, handling customer assignments and capacity constraints, is formulated as a Linear Program (LP). For this experiment, both the master QUBO and the LP subproblems are solved using Gurobi to provide a fair, head-to-head comparison of the algorithmic structure's efficiency against Method 1.

\end{itemize}

\subsubsection{The Scalability Bottleneck of Monolithic QUBO Formulations}
Before evaluating our hybrid method, we first demonstrate the fundamental challenge it is designed to overcome: the poor scalability of monolithic QUBO formulations. We applied the Monolithic QUBO Solver (Method 2) to a moderately sized CFLP instance (20 facilities, 20 customers) from the Discrete Location Problems Benchmark library \cite{beresnev2011}.

\begin{figure}[h] 
\centering
\includegraphics[width=1\linewidth]{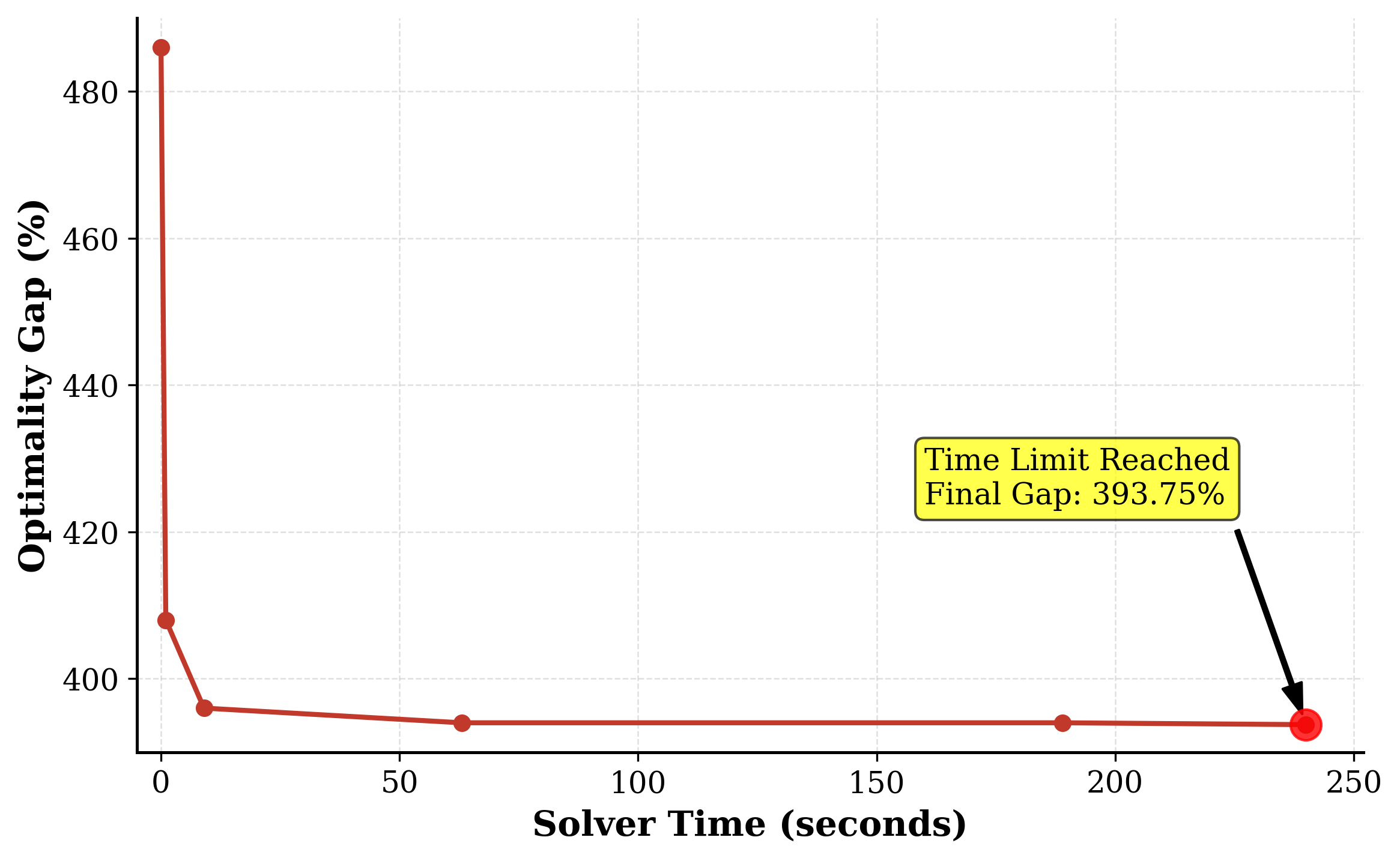} \caption{Convergence Failure of the Monolithic QUBO Approach on a 20x20 CFLP Instance.} 
\label{fig:your-label}
\end{figure}

Figure 2 illustrates the solver's performance in this task. The plot tracks the convergence of the optimality gap over time, showing that despite rapid initial improvements, the solver quickly stagnates. It does not reduce the gap below 393\% before reaching the 240-second time limit. This result underscores the intractability of solving non-trivial optimization problems via a direct, monolithic QUBO conversion; the enormous size and complex structure of the resulting matrix create a prohibitively difficult search landscape, thereby motivating the need for intelligent decomposition.

\subsubsection{Performance and Scalability of Hybrid Benders Decomposition}
We validate our framework on a suite of benchmark instances for the CFLP drawn from the well-established OR-Library, originally introduced by Ahuja et al. and Chen and Ting, and utilize the specific problem sets and their corresponding optimal solution values as benchmarked in Guastaroba and Speranza \cite{ahuja2004, chen2008, guastaroba2014}. In this section, we compare the performance of our Hybrid Benders Decomposition (Method 3) against the state-of-the-art Direct MILP Solver (Method 1) across these instances, with sizes scaling from 16 facilities and 50 customers up to 100 facilities and 1000 customers.

\begin{figure}[h] 
\centering
\includegraphics[width=1\linewidth]{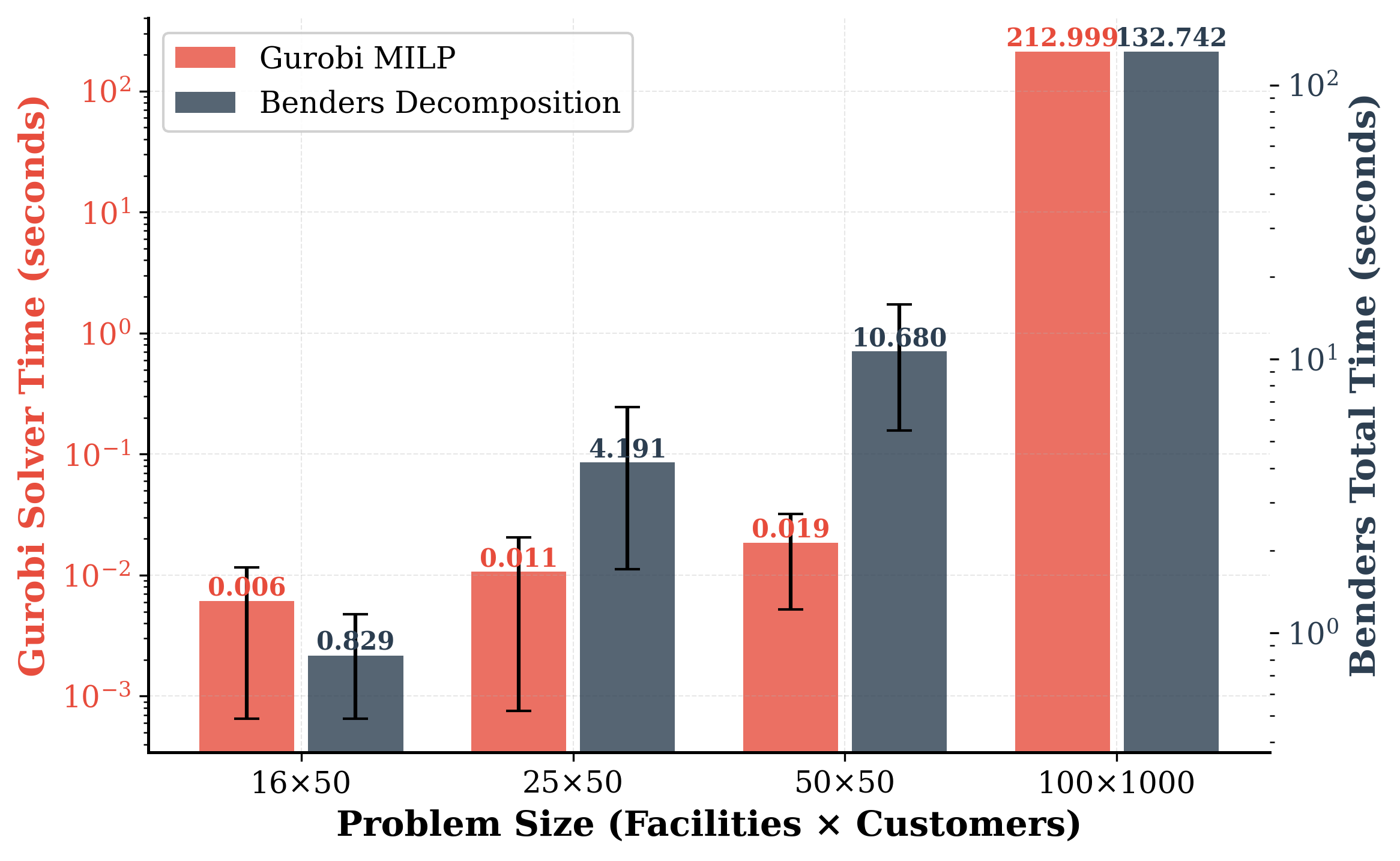} \caption{Scalability Comparison of the Hybrid Benders Decomposition against a Direct MILP Solver.} 
\label{fig:your-label}
\end{figure}

\begin{figure*}[h]
\centering
\includegraphics[width=0.9\linewidth]{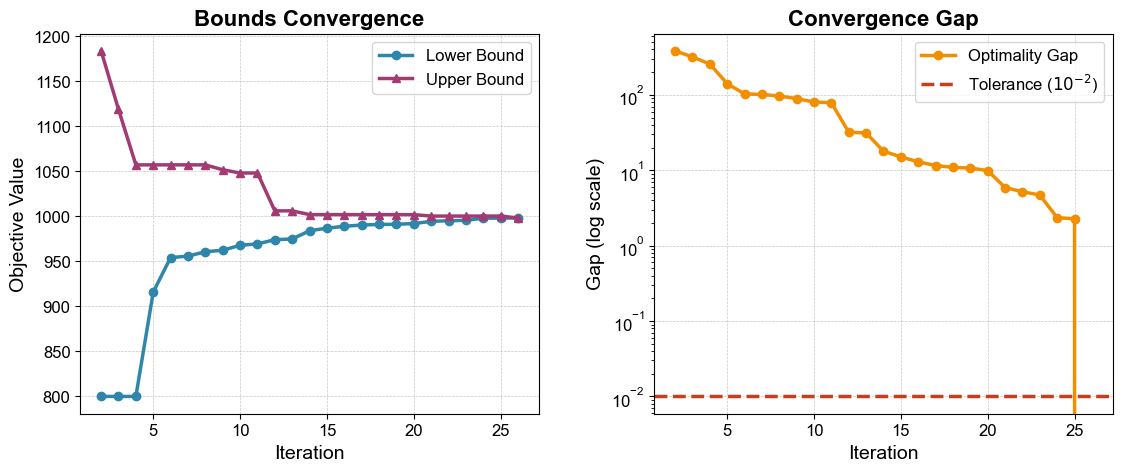}
\caption{Overview of the convergence behavior in the CFLP Benders decomposition approach}
\label{fig:convergence-overview}
\end{figure*}

\begin{figure}[ht] 
\centering
\includegraphics[width=1\linewidth]{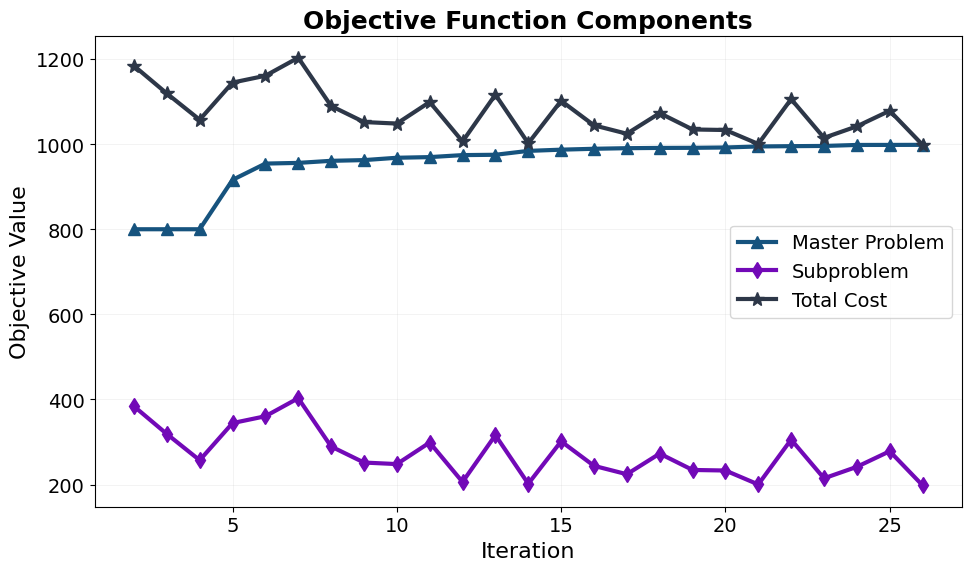} \caption{Evolution of Objective Function Components During Benders Decomposition.} 
\label{fig:your-label}
\end{figure}

Figure 3 presents the core results of our scalability analysis, comparing the performance of our Hybrid Benders Decomposition (Method 3, total time, gray) versus the Gurobi Direct MILP Solver (Method 1, solver time, red) on CFLP instances of increasing scale. Both y-axes are on a logarithmic scale, and error bars represent one standard deviation over multiple runs. For smaller instances (e.g., $16 \times 50$ and $25 \times 50$), the overhead of the Benders iterative loop makes it slower than the highly optimized direct Gurobi solver. However, the advantage of our decomposition approach becomes strikingly evident at the largest scale (100 facilities and 1000 customers). While the direct Gurobi solver required 213.0 seconds to find and prove the optimal solution, our hybrid framework obtained a high-quality solution with a negligible optimality gap of only 0.2\% in just 132.7 seconds, achieving a 38\% reduction in runtime compared to the direct Gurobi solver. It is crucial to note that these compelling results were obtained by solving the QUBO master problem classically. As the master problem is designed for a quantum annealer, we anticipate further performance gains when integrating a QPU backend, positioning our framework as a potent quantum accelerator for the combinatorial core of large-scale problems.

Figures 4 and 5 provide a detailed view into the internal mechanics of the Benders' decomposition process for a representative instance. Figure 4 demonstrates the characteristic convergence pattern, where the lower bound (derived from the master problem's relaxed solutions) monotonically increases while the upper bound (from feasible solutions found by the subproblem) decreases, robustly closing the optimality gap in under 30 iterations. Figure 5 dissects the objective value to illustrate the dynamics of the algorithm. The plot shows the Master Problem objective (which forms a monotonically increasing lower bound), the Subproblem cost (representing the optimal transportation cost for a given set of open facilities), and the Total Cost (the upper bound derived from the combined feasible solution in each iteration). This visualization highlights the interplay between the combinatorial master problem and the linear subproblem as they guide the search toward convergence.

\section{Conclusion and Future Work}
In this paper, we present a novel framework that leverages an LLM to automate the end-to-end pipeline from a standard MILP model to a QUBO formulation. Our experiments demonstrate that the proposed LLM-driven conversion process is both stable and capable of producing high-quality QUBO models that are semantically equivalent to their source problems. Furthermore, by integrating this core methodology into a Benders' decomposition scheme, we have demonstrated a powerful and scalable approach for solving large-scale optimization problems. This successful integration represents a significant step towards a new paradigm of hybrid quantum-AI computation.

For future work, we plan to evolve our methodology from its current rule-based implementation to a more robust learning-based approach. The current system, which relies on structured prompt engineering, offers the significant advantage of rapid iteration and validation of conversion strategies. Having established an effective set of principles through this method, our next step will be to use these curated data to fine-tune an LLM. By shifting from a context-based approach (providing rules in the prompt) to a learning-based one (embedding knowledge into the model's weights), we anticipate further enhancing the framework's robustness, improving its generalization capabilities across a wider range of problem classes, and potentially discovering even more efficient QUBO formulation strategies.

\bibliography{aaai24}

\begin{thebibliography}{23}
\providecommand{\natexlab}[1]{#1}

\bibitem[{Ahmed and Choudhury(2024)}]{ahmed2024}
Ahmed, T.; and Choudhury, S. 2024.
\newblock {{LM4OPT}}: {{Unveiling}} the Potential of {{Large Language Models}} in Formulating Mathematical Optimization Problems.
\newblock \emph{INFOR: Information Systems and Operational Research}, 62(4): 559--572.

\bibitem[{Ahuja et~al.(2004)Ahuja, Orlin, Pallottino, Scaparra, and Scutell{\`a}}]{ahuja2004}
Ahuja, R.~K.; Orlin, J.~B.; Pallottino, S.; Scaparra, M.~P.; and Scutell{\`a}, M.~G. 2004.
\newblock A {{Multi-Exchange Heuristic}} for the {{Single-Source Capacitated Facility Location Problem}}.
\newblock \emph{Management Science}, 50(6): 749--760.

\bibitem[{Ayodele(2022)}]{ayodele2022}
Ayodele, M. 2022.
\newblock Penalty {{Weights}} in {{QUBO Formulations}}: {{Permutation Problems}}.
\newblock In P{\'e}rez~C{\'a}ceres, L.; and Verel, S., eds., \emph{Evolutionary {{Computation}} in {{Combinatorial Optimization}}}, volume 13222, 159--174. Cham: Springer International Publishing.
\newblock ISBN 978-3-031-04147-1 978-3-031-04148-8.

\bibitem[{Beresnev et~al.(2011)Beresnev, Kochetov, Pashchenko, Goncharov, and Plyasunov}]{beresnev2011}
Beresnev, V.; Kochetov, Y.; Pashchenko, M.; Goncharov, E.; and Plyasunov, A. 2011.
\newblock Discrete {{Location Problems}} {{Benchmark}} Library.

\bibitem[{Chen and Ting(2008)}]{chen2008}
Chen, C.-H.; and Ting, C. 2008.
\newblock Combining {{Lagrangian}} Heuristic and {{Ant Colony System}} to Solve the {{Single Source Capacitated Facility Location Problem}}.
\newblock \emph{Transportation Research Part E-logistics and Transportation Review}, 44(6): 1099--1122.

\bibitem[{Chen et~al.(2025)Chen, {Constante-Flores}, Mantri, Kompalli, Ahluwalia, and Li}]{chen2025}
Chen, H.; {Constante-Flores}, G.~E.; Mantri, K. S.~I.; Kompalli, S.~M.; Ahluwalia, A.; and Li, C. 2025.
\newblock {{OptiChat}}: {{Bridging Optimization Models}} and {{Practitioners}} with {{Large Language Models}}.
\newblock \emph{ArXiv}, abs/2501.08406.

\bibitem[{Glover et~al.(2022)Glover, Kochenberger, Hennig, and Du}]{glover2022}
Glover, F.~W.; Kochenberger, G.~A.; Hennig, R.; and Du, Y. 2022.
\newblock Quantum Bridge Analytics {{I}}: A Tutorial on Formulating and Using {{QUBO}} Models.
\newblock \emph{Ann. Oper. Res.}, 314(1): 141--183.

\bibitem[{Guastaroba and Speranza(2014)}]{guastaroba2014}
Guastaroba, G.; and Speranza, M. 2014.
\newblock A Heuristic for {{BILP}} Problems: {{The Single Source Capacitated Facility Location Problem}}.
\newblock \emph{Eur. J. Oper. Res.}, 238(2): 438--450.

\bibitem[{Holliday(2025)}]{holliday2025}
Holliday, J. 2025.
\newblock Solving {{Real-World Optimization Problems}} Using {{Near-Term Quantum Computing}} with {{Applications}} in {{Vehicle Routing}} and {{Drone Delivery}}.
\newblock \emph{Graduate Theses and Dissertations}.

\bibitem[{Huang et~al.(2025)Huang, Tang, Hu, Jiang, Zheng, Ge, Wang, and Wang}]{huang2025}
Huang, C.; Tang, Z.; Hu, S.; Jiang, R.; Zheng, X.; Ge, D.; Wang, B.; and Wang, Z. 2025.
\newblock {{ORLM}}: {{A Customizable Framework}} in {{Training Large Models}} for {{Automated Optimization Modeling}}.
\newblock \emph{Operations Research}.

\bibitem[{Jiang et~al.(2025)Jiang, Shu, Qian, Lu, Zhou, Zhou, and Yu}]{jiang2025}
Jiang, C.; Shu, X.; Qian, H.; Lu, X.; Zhou, J.; Zhou, A.; and Yu, Y. 2025.
\newblock {{LLMOPT}}: {{Learning}} to {{Define}} and {{Solve General Optimization Problems}} from {{Scratch}}.
\newblock In \emph{Proceedings of the {{The Thirteenth International Conference}} on {{Learning Representations}}}, volume abs/2410.13213. OpenReview.net.

\bibitem[{Malviya, AkashNarayanan, and Seshadri(2023)}]{malviya2023}
Malviya, G.; AkashNarayanan, B.; and Seshadri, J. 2023.
\newblock Logistics {{Network Optimization Using Quantum Annealing}}.
\newblock In Noor, A.; Saroha, K.; Pricop, E.; Sen, A.; and Trivedi, G., eds., \emph{Proceedings of {{Third Emerging Trends}} and {{Technologies}} on {{Intelligent Systems}}}, 401--413. Singapore: Springer Nature.
\newblock ISBN 978-981-99-3963-3.

\bibitem[{Moraglio, Georgescu, and Sadowski(2022)}]{moraglio2022}
Moraglio, A.; Georgescu, S.; and Sadowski, P. 2022.
\newblock {{AutoQubo}}: Data-Driven Automatic {{QUBO}} Generation.
\newblock \emph{Proceedings of the Genetic and Evolutionary Computation Conference Companion}, 2232--2239.

\bibitem[{Morapakula et~al.(2025)Morapakula, Deshpande, Yata, Ubale, Wad, and Ikeda}]{morapakula2025}
Morapakula, S.~N.; Deshpande, S.; Yata, R.; Ubale, R.; Wad, U.; and Ikeda, K. 2025.
\newblock End-to-{{End Portfolio Optimization}} with {{Quantum Annealing}}.

\bibitem[{M{\"u}cke, Gerlach, and Piatkowski(2023)}]{mucke2023}
M{\"u}cke, S.; Gerlach, T.; and Piatkowski, N. 2023.
\newblock Optimum-Preserving {{QUBO}} Parameter Compression.
\newblock \emph{Quantum Mach. Intell.}, 7: 1.

\bibitem[{Oliveira, Silva, and Oliveira(2018)}]{oliveira2018}
Oliveira, N. M.~D.; Silva, R. M. D.~A.; and Oliveira, W. R.~D. 2018.
\newblock {{QUBO}} Formulation for the Contact Map Overlap Problem.
\newblock \emph{International Journal of Quantum Information}, 16(08): 1840007.

\bibitem[{Pauckert et~al.(2023)Pauckert, Ayodele, Garc{\'i}a, Georgescu, and Parizy}]{pauckert2023}
Pauckert, J.; Ayodele, M.; Garc{\'i}a, M.; Georgescu, S.; and Parizy, M. 2023.
\newblock {{AutoQUBO}} v2: {{Towards Efficient}} and {{Effective QUBO Formulations}} for {{Ising Machines}}.
\newblock \emph{Proceedings of the Companion Conference on Genetic and Evolutionary Computation}, 227--230.

\bibitem[{Ramamonjison et~al.(2022)Ramamonjison, Li, Yu, He, Rengan, {Banitalebi-Dehkordi}, Zhou, and Zhang}]{ramamonjison2022}
Ramamonjison, R.; Li, H.; Yu, T.~T.; He, S.; Rengan, V.; {Banitalebi-Dehkordi}, A.; Zhou, Z.; and Zhang, Y. 2022.
\newblock Augmenting {{Operations Research}} with {{Auto-Formulation}} of {{Optimization Models}} from {{Problem Descriptions}}.

\bibitem[{Ramamonjison et~al.(2021)Ramamonjison, Yu, Li, Li, Carenini, Ghaddar, He, Mostajabdaveh, {Banitalebi\{-\}Dehkordi}, Zhou, and Zhang}]{ramamonjison2021}
Ramamonjison, R.; Yu, T. T.~L.; Li, R.; Li, H.; Carenini, G.; Ghaddar, B.; He, S.; Mostajabdaveh, M.; {Banitalebi\{-\}Dehkordi}, A.; Zhou, Z.; and Zhang, Y. 2021.
\newblock {{NL4Opt Competition}}: {{Formulating Optimization Problems Based}} on {{Their Natural Language Descriptions}}.
\newblock In \emph{Proceedings of the {{NeurIPS}} 2022 {{Competition Track}}}, volume 220 of \emph{Proceedings of {{Machine Learning Research}}}, 189--203. PMLR.

\bibitem[{Volpe et~al.(2024)Volpe, Quetschlich, Graziano, Turvani, and Wille}]{volpe2024}
Volpe, D.; Quetschlich, N.; Graziano, M.; Turvani, G.; and Wille, R. 2024.
\newblock Towards an {{Automatic Framework}} for {{Solving Optimization Problems}} with {{Quantum Computers}}.
\newblock In \emph{2024 {{IEEE International Conference}} on {{Quantum Software}} ({{QSW}})}, 46--57.

\bibitem[{Zaman, Tanahashi, and Tanaka(2022)}]{zaman2022}
Zaman, M.; Tanahashi, K.; and Tanaka, S. 2022.
\newblock {{PyQUBO}}: {{Python Library}} for {{Mapping Combinatorial Optimization Problems}} to {{QUBO Form}}.
\newblock \emph{IEEE Trans. Computers}, 71: 838--850.

\bibitem[{Zhao, Fan, and Han(2022)}]{zhao2022}
Zhao, Z.; Fan, L.; and Han, Z. 2022.
\newblock Hybrid {{Quantum Benders}}' {{Decomposition For Mixed-integer Linear Programming}}.
\newblock \emph{2022 IEEE Wireless Communications and Networking Conference (WCNC)}, 2536--2540.

\bibitem[{Zhao et~al.(2025)Zhao, Li, Fan, and Han}]{zhao2025}
Zhao, Z.; Li, M.; Fan, L.; and Han, Z. 2025.
\newblock {{HQC-Bend}}: {{A Python Package}} of {{Hybrid Quantum-Classical Multi-cuts Benders}}' {{Decomposition Algorithm}}.
\newblock \emph{2025 International Conference on Quantum Communications, Networking, and Computing (QCNC)}, 591--597.

\end{thebibliography}

\end{document}